\documentclass[review]{elsarticle}

\usepackage{multirow}
\usepackage{amssymb}
\usepackage{amsmath}
\usepackage{lineno}
\usepackage{subcaption}
\usepackage{graphicx}
\usepackage{url}

\begin{document}
	
	\title{A self-supervised learning approach to deep filter banks for texture recognition}
	
	\author[2]{Florindo, Joao B.}
	\author[1]{Lyra, Lucas O.}
	\author[1]{Fabris, Antonio E.}
	
	\affiliation[1]{organization={Institute of Mathematics and Statistics of the University of Sao Paulo},
		addressline={Rua do Matao, 1010}, 
		city={Sao Paulo},
		postcode={05508-090}, 
		state={Sao Paulo},
		country={Brazil}}
	
	\affiliation[2]{organization={Institute of Mathematics, Statistics and Scientific Computing of the University of Campinas},            
		addressline={Rua Sergio Buarque de Holanda, 651}, 
		city={Campinas},
		postcode={13083-859}, 
		state={Sao Paulo},
		country={Brazil}}	
	
	\begin{abstract}
		An important challenge in texture recognition is the limited amount of data for training frequently found in real-world applications. In computer vision in general, a successful strategy to mitigate this issue is the use of a pretraining stage where the neural network learns to identify relations between parts of the data in a self-supervised manner. A well-established framework in this direction is masked autoencoder. Nevertheless, these models usually rely on computationally intensive architectures, such as vision transformers. In the particular case of texture images, most of the relevant information is compacted within a delimited area around each pixel, which suggests that capturing long-range dependence via the attention mechanism may be unnecessary. Based on that assumption, here we propose a framework where the pretraining model is a convolutional autoencoder. To leverage the rich information conveyed by texture patterns, we employ deep filters coupled with Fisher vector pooling. In this way, we improve the performance of texture recognition without adding significant computational burden. Our approach is compared with several state-of-the-art methods in different texture databases, confirming its potential both in terms of classification accuracy and computational complexity.
	\end{abstract}
	
	
	\maketitle
	
	
	\section{Introduction}
	
	Texture recognition is a fundamental task in computer vision, with applications to diverse areas, such as in medicine \cite{young2024diagnosis}, agriculture \cite{barburiceanu2021convolutional}, material sciences \cite{si2024v}, remote sensing \cite{han2024remote}, and many others.
	
	Despite the efforts in the literature to improve the performance of deep learning methods in this problem, we still have some challenges in most real-world applications of texture analysis, e.g., the usually reduced amount of data for training, high interclass similarity and intraclass dissimilarity. In this scenario, the use ``off-the-shelf'' of convolutional neural networks (CNN) does not achieve optimal results. Most texture recognition methodologies rely on extracting more information from the training data than a CNN can typically represent. And a modern strategy that has been successful in situations like that is the use of a self-supervised pretraining stage \cite{akiva2022self}.
	
	Self-supervised algorithms in computer vision can be usually grouped into contrastive, clustering-based, and autoassociative learning. The last category, in particular, is usually more effective in feature extraction, which is a crucial step in texture recognition algorithms. A prototypical representer of autoassociative model is masked auto-encoder (MAE) \cite{he2022masked}. Its pretraining task consists of reconstructing images with randomly masked patches. Such reconstruction requires the understanding of long-range relations within the image, which motivates the use of Vision Transforms (ViT) \cite{dosovitskiy2020image} instead of convolutional networks. On the other hand, using ViT adds significant computational burden to the pretraining stage and useful deep representations of texture images are known for a long time to be highly dependent on local characteristics.
	
	Based on that, here we propose a convolutional auto-encoder for the pretraining stage of texture recognition tasks. Two approaches are explored: 1) The input and output of the autoencoder are the same image; 2) The input corresponds to a noisy version of the original image, which should be recovered at the output. In summary, the main contributions of this manuscript are:
	
	\begin{itemize}
		\item We propose a deep convolutional encoder-decoder architecture that extracts meaningful features from texture images in a self-supervised manner.
		\item A model for texture recognition is introduced, using a combination of self-supervised CNN and Fisher Vectors.
		\item We advance the current state-of-the-art results on the Flickr Material Database (FMD)~\cite{sharan2014accuracy}, the Describable Textures Dataset (DTD)~\cite{cimpoi2014describing}, and the KTH-TIPS2-b image database~\cite{caputo2005class}.
		\item The proposed model demonstrates outstanding accuracy in the practical task of identifying Brazilian plant species, substantially surpassing previously reported results in the literature.
	\end{itemize}

	Section~\ref{sec:related} reviews the literature relevant to our research.  
	In Section~\ref{sec:background}, we outline the theoretical foundations necessary for understanding our method.  
	Section~\ref{sec:proposed} provides a detailed description of our proposed approach for medical image classification.  
	The experimental setup, including dataset descriptions and implementation specifics, is presented in Section~\ref{sec:experiments}.  
	Section~\ref{sec:results} presents and analyzes our experimental findings.  
	Finally, Section~\ref{sec:conclusions} offers concluding remarks and suggestions for future research directions.
	
	\section{Related works}
	\label{sec:related}
	
	In this section, we provide a brief overview of studies relevant to our work. We begin by examining the literature on autoencoders, then discuss orderless encoding methods, and finally review general approaches to texture recognition.
	
	\subsection{Autoencoders}
	
	Despite being an algorithm typically associated with unsupervised learning, we can find examples of the use of autoencoders on image recognition.
	
	In \cite{gogna2019discriminative}, an autoencoder is employed for representation learning. The extracted features provide interesting results in character recognition even when applied to simple classifiers like nearest neighbors. The authors in \cite{yang2022orthogonal} propose the concept of \textit{orthogonal autoencoder regression}, where the representation capability of autoencoders is combined with least square regression for image classification. Autoencoders are also frequently used as an auxiliary tool to improve the classification pipeline in general. For example, in \cite{kang2023deblurring}, a masked autoencoder is used for debluring of ultrasound images.
	
	\subsection{Orderless Encoding}
	
	In visual texture classification, the scarcity of labeled data and the domain gap from ImageNet have motivated many researchers to leverage pre-trained CNN weights. Orderless encoding methods are particularly well-suited for this field and are widely adopted. Cimpoi et al.~\cite{cimpoi2015deep} introduced the use of Fisher Vectors to encode features extracted from the final convolutional layer of a VGG network~\cite{simonyan2014very}. Building on this idea, Lyra et al.~\cite{lyra2024multilevel} developed a novel approach for extracting features from multiple convolutional layers across various CNN architectures, achieving improved accuracy compared to using CNNs alone.
	
	Beyond Fisher Vectors, several other orderless encoding techniques have been explored. Chen et al.~\cite{chen2021deep} also extract features from multiple convolutional layers, but concatenate the feature maps by upsampling lower-resolution maps using bilinear interpolation. They then assess statistical self-similarity with a differential box counting method and compute soft histograms, which are concatenated with average pooling to form the final image descriptor. RADAM~\cite{scabini2023radam} employs a similar cross-layer concatenation strategy, but generates the image descriptor using a randomized autoencoder. In~\cite{yang2022dfaen}, first- and second-order statistics are extracted from feature maps using a frequency attention mechanism and encoded with bilinear models. Xu et al.~\cite{xu2021encoding} propose combining fractal average pooling with global average pooling to create more robust descriptors. Florindo et al.~\cite{florindo2023boff} compute descriptors using fuzzy equivalence measures applied to clustered local features.
	
	\subsection{Texture Recognition}
	
	Recent advances in texture recognition have been driven by developments in deep learning architectures, feature extraction strategies, and hybrid modeling approaches.
	
	The introduction of Vision Transformers (ViTs) has brought new perspectives to texture recognition. Scabini et al.~\cite{scabini2024comparative} conducted an extensive evaluation of 21 ViT variants, including ViT-B with DINO pre-training, BeiT v2, and Swin Transformers. Their findings indicate that ViTs generally surpass traditional CNNs and hand-crafted models, especially for in-the-wild texture tasks. EfficientFormer was identified as a cost-effective alternative, delivering strong performance with lower computational requirements. Scabini et al.~\cite{scabini2023radam} also proposed RADAM (Randomized Aggregated Deep Activation Maps), a method that encodes texture representations without fine-tuning the backbone. RADAM uses a Randomized Autoencoder (RAE) trained locally on each image to process outputs from various depths of a pre-trained CNN, producing a 1D texture representation classified by a linear SVM. This approach achieves state-of-the-art results on multiple texture benchmarks, demonstrating the power of pre-trained backbones without additional fine-tuning.
	
	To address CNNs' limitations in capturing fine local details, Zhu et al.~\cite{zhu2023learning} integrated Gabor filters into CNN architectures. Their method introduces a texture branch that extracts multi-frequency features using Gabor filters, complemented by a statistical feature extractor and a gate selection mechanism. This hybrid design enhances recognition of fine-grained categories, achieving state-of-the-art results on datasets such as CUB-200-2011 and GTOS-mobile.
	
	Bera et al.~\cite{bera2022deep} proposed a fusion approach that combines global texture features with local patch-based information for fine-grained image classification. Their method extracts deep features from fixed-size, non-overlapping patches, encodes them with LSTM, and computes image-level textures using Local Binary Patterns (LBP). The integration of these streams yields an efficient feature vector, leading to improved classification accuracy across diverse datasets, including those for human faces and skin lesions.
	
	Goyal et al.~\cite{goyal2023texture} investigated transfer learning for texture classification using pre-trained models such as MobileNetV3 and InceptionV3. Their experiments on datasets like Brodatz, Kylberg, and Outex demonstrated high classification accuracy, underscoring the effectiveness of transfer learning in texture analysis.
	
	\section{Background}
	\label{sec:background}
	
	In this section, we outline two fundamental principles underlying our methodology. The first one is the autoencoder, which handles feature extraction. The second is Fisher Vector encoding, which generates the feature vectors used for classification.
	
	\subsection{Autoencoder}
	
	The encoder mathematically performs a series of operations:
	\begin{align}\label{eq:ae1}
	\begin{split}
	&E_i(x)=\sigma(W_{i,2}\ast \sigma(W_{i,1}\ast E_{i-1}(x)+b_{i,1})+b_{i,2})\\
	&D_i(x)=P(E_i(x)),
	\end{split}
	\end{align}
	where $E_i$ represents the output of the $i$-th encoder block, $D_i$ represents the downsampled output after pooling, $W_{i,j}$ are the convolution kernels ($3\times 3$), $b_{i,j}$ are bias terms, $\sigma$ is the ReLU activation function, $P$ is the max pooling operation ($2\times 2$ with stride 2).
	
	The bottleneck connects the encoder and decoder:
	\begin{equation}
	\label{eq:ae2}
	B(x)=\sigma(W_{b,2}\ast \sigma(W_{b,1}\ast D_J(x)+b_{b,1})+b_{b,2}),
	\end{equation}
	where $J$ is the final encoder level and $B$ is the bottleneck output.
	
	The decoder performs upsampling followed by convolutions:
	\begin{align}
	\begin{split}
	&U_i(x)=W_{u,i}\ast U(D_{J-i+1}(x))\\
	&C_i(x)=concat(E_{J-i+1}(x),U_i(x))\\
	&F_i(x)=\sigma(W_{d,i,2}\ast \sigma(W_{d,i,1}\ast C_i(x)+b_{d,i,1})+b_{d,i,2}),
	\end{split}
	\end{align}
	where $U$ is the upsampling operation, $W_{u,i}$ are the transposed convolution weights, $C_i$ is the concatenation of encoder features with upsampled decoder features, and $F_i$ is the output of the $i$-th decoder block.
	
	The skip connections can be mathematically represented as:
	\begin{equation}\label{eq:ae4}
	C_i(x)=concat(E_{J-i+1}(x),U_i(x)).
	\end{equation}
	These connections preserve spatial information that would otherwise be lost during encoding, enhancing the precision of segmentation outputs
	
	\subsection{Fisher Vector}
	\label{sec:fisher}
	
	Let \( X = \{\mathbf{x}_t \in \mathbb{R}^D \mid t = 1, \ldots, T\} \) represent a sample containing \( T \) observations. Assuming the generation process of \( X \) follows a probability density function \( u_\lambda \) with parameters \( \lambda \), we characterize these observations through the gradient vector:
	
	\begin{equation}
		\label{eq:grad}
		G_{\lambda}^X = \nabla_{\lambda} \log u_{\lambda}(X).
	\end{equation}
	
	This gradient vector from Equation~\ref{eq:grad} can be employed in classification tasks. Following \cite{jaakkola1998exploiting}, we use the Fisher information matrix \( F_{\lambda} \) defined as:
	\begin{equation}
		F_{\lambda} = \mathbb{E}_X\left[G_{\lambda}^X (G_{\lambda}^X)^\top\right],
	\end{equation}
	where \( \mathbb{E}_X \) denotes the expectation over \( X \).
	
	The Fisher Kernel (FK) measuring similarity between two samples \( X \) and \( Y \) is then given by:
	\begin{equation}
		K_{\text{FK}}(X,Y) = (G_{\lambda}^X)^\top F_{\lambda}^{-1} G_{\lambda}^Y.
	\end{equation}
	
	Since \( F_{\lambda}^{-1} \) is positive semi-definite, we apply Cholesky decomposition \( F_{\lambda}^{-1} = L_{\lambda}^\top L_{\lambda} \) to reformulate the kernel as:
	\begin{equation}
		K_{\text{FK}}(X,Y) = (\mathcal{G}_{\lambda}^X)^\top \mathcal{G}_{\lambda}^Y,
	\end{equation}
	where the Fisher Vector (FV) is defined by:
	\begin{equation}
		\label{eq:fisher}
		\mathcal{G}_{\lambda}^X = L_{\lambda} G_{\lambda}^X.
	\end{equation}
	
	Notably, the Fisher Vector \( \mathcal{G}_{\lambda}^X \) maintains the same dimensionality as \( G_{\lambda}^X \) \cite{sanchez2013image}. This implies that classification using a linear kernel machine with FVs is equivalent to employing a non-linear kernel machine with \( K_{\text{FK}} \). For comprehensive details on Fisher Vectors, we refer readers to \cite{sanchez2013image}.
	
	\section{Proposed Method}
	\label{sec:proposed}
	
	The proposed methodology can be organized into 4 major modules: autoencoder self-supervised module, CNN feature extractor, GMM training, and predictor. The following sections delineate the overall idea of each such component. 
	
	\subsection{Self-supervised module}
	
	The self-supervised module works at two stages of the pipeline. In the pretraining task, the input image is processed by Equations (\ref{eq:ae1})-(\ref{eq:ae4}). 
	
	Being an autoencoder, we have that both the input and target output correspond to the same training image $I \in \mathbb{R}^{m\times n}$. For the reconstruction loss we adopt the classical mean squared error, such that all weights $W$ and biases $b$ in (\ref{eq:ae1})-(\ref{eq:ae4}) are learned by minimizing
	\[ 
		L = \frac{1}{mn}\sum_{x=1}^{m}\sum_{y=1}^{n}\left( I[x,y]-F_i(I)[x,y] \right)^2,
	\]
	where $F_i$ is defined in (\ref{eq:ae4}).
	
	In the final classification task, the decoder is removed and the output is provided by $B(I)$, where $B$ is the bottleneck operator defined by Equation (\ref{eq:ae2}).
	
	\subsection{CNN Feature Extractor}
	
	This module consists of a series of convolutional operations. At each layer $\ell$, for an input with $C$ channels, the output at position $(i, j)$ and channel $k$ is defined as
	\[
	x_{ijk\ell} = \sum_{c=1}^C \sum_{a=0}^{m-1} \sum_{b=0}^{m-1} w_{abck} \cdot y^{\ell-1}_{(i+a)(j+b)c},
	\]
	where $m$ is the kernel size, $w$ represents the (potentially learnable) weights, and
	\[
	y_{ijk}^{\ell} = \sigma(x_{ijk}^{\ell}),
	\]
	with $\sigma$ denoting a nonlinear activation function. This architecture may also include other standard operations, such as pooling, depending on the definition of $w$.
	
	\subsection{GMM Training}
	
	Fisher vectors are computed from the aggregated set of local features $X = \{\mathbf{x}_t \in \mathbb{R}^D \mid t = 1, 2, \ldots, T\}$, where $T = \sum_{i=0}^{k} T_{n-i}$ and $D = D_{n-k}$.
	
	We assume each local feature $\mathbf{x}_t$ is independently generated according to the distribution $u_{\lambda}$. Under this assumption, Equation~(\ref{eq:fisher}) becomes:
	\begin{equation}
		\mathcal{G}_{\lambda}^{X} = L_{\lambda} \frac{1}{T} \sum_{t=1}^{T} \nabla_{\lambda} \log u_{\lambda}(\mathbf{x}_t).
	\end{equation}
	
	Here, $u_{\lambda}$ is modeled as a Gaussian Mixture Model (GMM) with $K$ Gaussian components, each representing a visual word in the learned dictionary:
	\begin{equation}
		u_{\lambda}(\mathbf{x}) = \sum_{i=1}^{K} w_i u_i(\mathbf{x}),
	\end{equation}
	where $\lambda = \{w_i, \mu_i, \Sigma_i \mid i = 1, \ldots, K\}$, and $w_i$, $\mu_i$, and $\Sigma_i$ denote the weight, mean, and covariance matrix of the $i$-th Gaussian component, respectively.
	
	The probability that an observation $\mathbf{x}_t$ is generated by the $i$-th Gaussian is given by:
	\begin{equation}
		\gamma_i(\mathbf{x}_t) = \frac{w_i u_i(\mathbf{x}_t)}{\sum_{j=1}^{K} w_j u_j(\mathbf{x}_t)}.
	\end{equation}
	
	We assume diagonal covariance matrices, as any distribution can be approximated to arbitrary precision by a weighted sum of Gaussians with diagonal covariances~\cite{perronnin2007fisher}. Let $\sigma_i^2 = \mathrm{diag}(\Sigma_i)$. Using the expressions for $L_\lambda$ and $\nabla_{\lambda} \log u_{\lambda}(X)$ from~\cite{perronnin2007fisher}, Equation~(\ref{eq:fisher}) can be rewritten as:
	\begin{align}
		\label{eq:fisher1}
		\mathcal{G}_{w_i^d}^{X} &= \frac{1}{T\sqrt{w_i}} \sum_{t=1}^{T} \left( \gamma_i(\mathbf{x}_t) - w_i \right), \\
		\label{eq:fisher2}
		\mathcal{G}_{\mu_i^d}^{X} &= \frac{1}{T\sqrt{w_i}} \sum_{t=1}^{T} \gamma_i(\mathbf{x}_t) \left( \frac{x_t^d - \mu_i^d}{\sigma_i^d} \right), \\
		\label{eq:fisher3}
		\mathcal{G}_{\sigma_i^d}^{X} &= \frac{1}{T\sqrt{2w_i}} \sum_{t=1}^{T} \gamma_i(\mathbf{x}_t) \left[ \frac{(x_t^d - \mu_i^d)^2}{(\sigma_i^d)^2} - 1 \right].
	\end{align}
	
	These equations (\ref{eq:fisher1}--\ref{eq:fisher3}) are used to compute each component of the Fisher Vector (FV) from any set of local features extracted from the CNN’s convolutional layers. Additionally, we extract features from the last fully connected layer by removing the output layer; this feature vector is referred to as FC.
	
	\subsection{Predictor}
	\label{sec:predictor}
	
	For classification, we apply both L2 and power normalization to the FV features, as recommended in~\cite{perronnin2010improving}. While various normalization techniques exist for different image types, some can be computationally expensive. Therefore, we adopt an efficient normalization strategy to ensure high performance.
	
	We further enhance the feature representation by concatenating the self-supervised component (SSL) with the normalized FV features, resulting in a combined feature vector denoted as SSL+FV.
	
	Classification is performed using a Support Vector Machine (SVM) with the Bhattacharyya coefficient, as defined in Definition~\ref{def:modified}, serving as the kernel. For $\mathbf{x}, \mathbf{y} \in \mathbb{R}^N$, the Bhattacharyya coefficient is defined as:
	\begin{equation}
		K(\mathbf{x}, \mathbf{y}) = \sum_{i=1}^{N} \mathrm{sign}(x_i y_i) \sqrt{|x_i y_i|}.
	\end{equation}
	This coefficient can be rewritten as:
	\begin{equation}
		K(\mathbf{x}, \mathbf{y}) = \phi(\mathbf{x})^{T} \phi(\mathbf{y}),
	\end{equation}
	where $\phi(\mathbf{x})$ is a vector with components:
	\begin{equation}
		\label{eq:phi}
		\phi(\mathbf{x})_i = \mathrm{sign}(x_i) \sqrt{|x_i|}.
	\end{equation}
	By applying the transformation in Equation~(\ref{eq:phi}) to the feature vectors, classification can be performed using a linear SVM. For multi-class problems, we employ a one-vs-rest strategy.
	
\begin{figure}[!htpb]
		\centering
		\includegraphics[width=\textwidth]{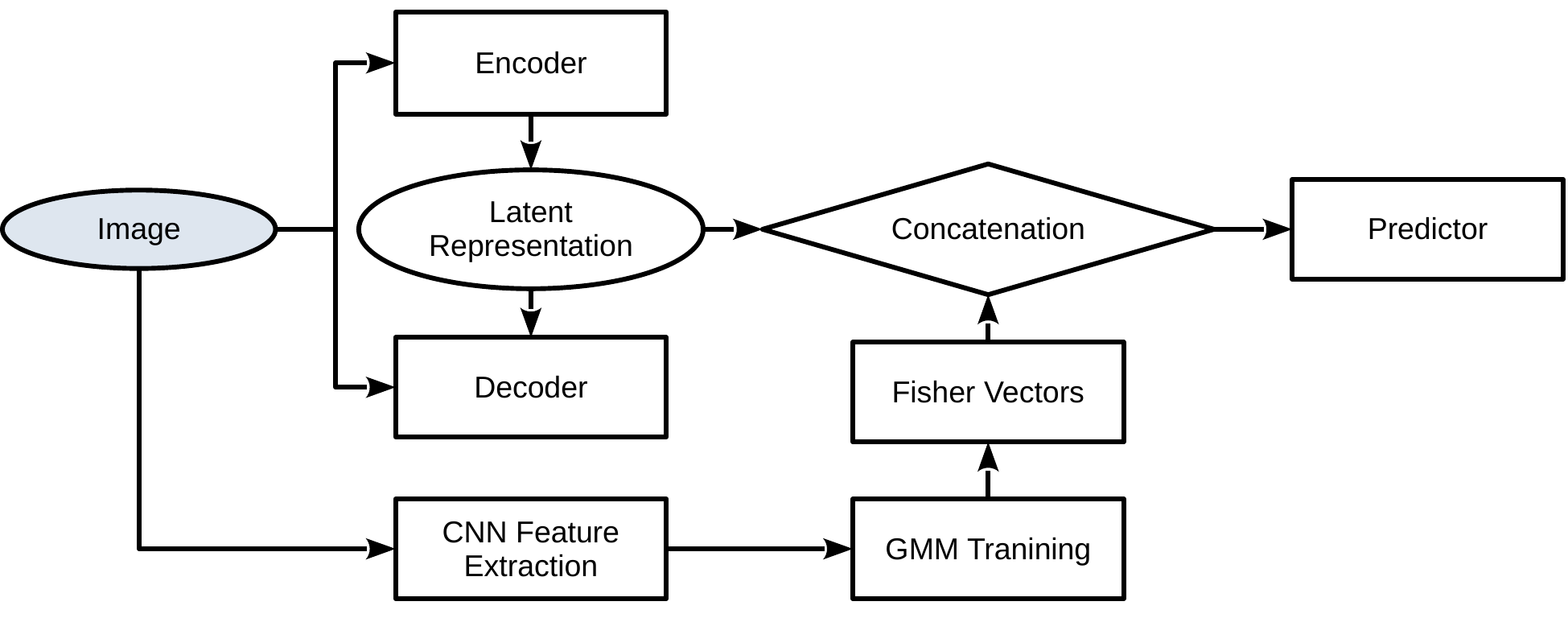}
		\caption{Proposed method.}
		\label{fig:method}
	\end{figure}

	Figure~\ref{fig:method} illustrates the overall architecture of the proposed framework.
		
	\section{Experimental setup}
	\label{sec:experiments}
	
	In this section, we outline the evaluation protocol for our proposed methodology. Our base model employs the EfficientNet-B5 architecture~\cite{tan2019efficientnet}, utilizing ImageNet pre-trained weights as the feature extractor.
	
	The datasets used for evaluation include KTH-TIPS2-b, FMD, DTD, UIUC, and UMD. For the practical application, we use the 1200Tex dataset. Each of these datasets is described below.
	
	KTH-TIPS2-b~\cite{caputo2005class} contains images of 11 materials, each with 4 samples. Every sample is captured at 9 scales, 3 poses, and 4 lighting conditions, resulting in 108 images per material per sample. For each evaluation round, 3 samples are used for training and 1 for testing.
	
	FMD~\cite{sharan2014accuracy} comprises 10 classes with 100 images each, at a resolution of 512$\times$384 pixels. We conduct 10 rounds of training and testing, randomly splitting the dataset in half for each round.
	
	DTD~\cite{cimpoi2014describing} consists of 5,640 images of varying sizes, divided into 47 categories (120 images per class). Each class is split equally into training, validation, and testing sets. The dataset provides 10 predefined splits; for each, we use the training and validation sets for model adjustment and the test set for evaluation.
	
	UMD~\cite{xu2009viewpoint} includes 25 classes with 40 images each, all sized 1280$\times$960 pixels. We follow the same evaluation protocol as for FMD.
	
	UIUC~\cite{lazebnik2005sparse} contains 1,000 images evenly distributed across 25 classes, each with a resolution of 640$\times$480 pixels. The evaluation protocol matches that used for FMD.
	
	1200Tex~\cite{CMB09} comprises 1,200 leaf surface images from 20 Brazilian plant species, with 60 samples per class. The same training and testing protocol as FMD is applied.
	
	To ensure fair comparison across datasets, we standardize the image width to yield a similar number of local features. For datasets with varying image sizes, both width and height are set to the same value, with a default width of $320$ pixels.
	
	Unless otherwise noted, Fisher Vectors are computed using 16 kernels. Local features are extracted from two convolutional layers: the last layer of block 5 and the last layer of block 6. Since the latter has more channels, dimensionality reduction is applied to its features, typically using PCA.
	
	Our experiments begin by comparing PCA with two alternative dimensionality reduction methods: average pooling and max pooling. Next, we assess how accuracy changes as the number of layers used for local feature extraction increases, including results for a single layer (the last convolutional layer), as in~\cite{cimpoi2016deep}. Layers are selected as the last convolutional layer of each block, starting from the final block; for $n$ layers, the last $n$ blocks are used.
	
	We also investigate how the number of kernels and the number of local features affect the descriptive power of Fisher Vectors. In the third experiment, we vary the number of kernels and evaluate the resulting accuracy. Subsequently, we examine the impact of the number of local features by adjusting the input image width.
	
	Finally, we compare our base model to alternative state-of-the-art methods, using optimized parameters for our approach. The experimental section concludes with a practical application: identifying Brazilian plant species from scanned images of leaf surfaces.
	
	\section{Results and Discussion}
	\label{sec:results}
	
	Table \ref{tab:metrics} lists different classification metrics for the proposed method on the analyzed datasets. Overall, we observe that using only Fisher features provide slightly better metrics. This can be explained by the intrinsic similarity in nature between FC features and AE features, as both are generated by a learnable CNN. In this regard, the statistical procedure supported by GMM is more effective as it takes into account particularities of the distribution of deep features across the dataset.
	\begin{table}[!htpb]
	\centering
	\caption{Classification metrics of the proposed method, in its two variations FCFVAE and FVAE.}
	\label{tab:metrics}
	\begin{tabular}{lcccc}
	\hline
	\multicolumn{5}{c}{FMD}\\
	\hline
	Feature Vector & Accuracy    & Precision   & Recall      & F1-Score   \\
	FV             & 90.3 $\pm$ 0.5 & 90.3 $\pm$ 0.5 & 90.5 $\pm$ 0.5 & 90.2 $\pm$ 0.5 \\
	FV+FC          & 91.7 $\pm$ 0.4 & 91.7 $\pm$ 0.3 & 91.8 $\pm$ 0.4 & 91.7 $\pm$ 0.4 \\
	FCFVAE      & 92.2 $\pm$ 0.2 & 92.3 $\pm$ 0.2 & 92.3 $\pm$ 0.2 & 92.2 $\pm$ 0.2 \\
	FVAE         & 92.9 $\pm$ 0.4 & 93.0 $\pm$ 0.4 & 93.0 $\pm$ 0.5 & 92.9 $\pm$ 0.4 \\
	\hline
	\multicolumn{5}{c}{KTH}\\
	\hline
	Feature Vector & Accuracy    & Precision   & Recall      & F1-Score   \\
	\hline
	FV             & 92.8 $\pm$ 4.9 & 93.9 $\pm$ 4.1 & 92.8 $\pm$ 4.9 & 92.2 $\pm$ 5.6 \\
	FV+FC          & 92.4 $\pm$ 4.6 & 93.5 $\pm$ 4.0 & 92.4 $\pm$ 4.6 & 92.0 $\pm$ 5.1 \\
	FCFVAE      & 92.3 $\pm$ 4.6 & 93.4 $\pm$ 4.0 & 92.3 $\pm$ 4.6 & 91.8 $\pm$ 5.1 \\
	FVAE         & 92.7 $\pm$ 4.8 & 93.7 $\pm$ 4.1 & 92.7 $\pm$ 4.8 & 92.2 $\pm$ 5.4 \\
	\hline
	\end{tabular}
	\end{table}
	
	Table \ref{tab:result} presents the accuracy achieved by the proposed method in comparison with several classical and state-of-the-art approaches in the literature. The overall performances of FVAE and FCFVAE show that our proposals are competitive with the state-of-the-art and represent a promising direction of exploration in texture recognition. Particularly, compared with the baseline backbone (EfficientNet-B5) of our architecture, we notice that FV+AE features contributed significantly, increasing the average accuracy in more than 5\% in most databases. Such numbers can be explained by the ability of autoencoders to compress information while preserving its essence. In a self-supervised framework, where the pre-training stage should be able of enriching the CNN with specific internal patterns of the image, this approach represents a straightforward and, at the same time, highly efficient way of obtaining such representation. 
	\begin{table}[!htpb]
		\caption{Accuracy comparison with other methods in literature. In the first row, we include the performance of fine-tuning the same CNN architecture we are using. All results shown are obtained directly from the original paper of each method. Non-published results are represented by dashes.}
		\label{tab:result}
		\centering
		\scalebox{.8}{
			\begin{tabular}{l c c c c c c}
				\hline
				Method & KTH-TIPS2-b & FMD & DTD & UMD & UIUC & GTOS \\
				\hline
				EfficientNet-B5 & $87.0_{\pm5.9}$ & $87.4_{\pm0.6}$ & $77.6_{\pm0.4}$ & $99.9_{\pm0.1}$ & $98.4_{\pm0.4}$ & $78.7_{\pm2.0}$\\
				\hline
				FV-VGGVD \cite{cimpoi2016deep} & $81.8_{\pm2.4}$ & $79.8_{\pm1.8}$ & $72.3_{\pm1.0}$ & $99.9_{\pm0.1}$ & $99.9_{\pm0.1}$ & - \\
				SIFT-FV \cite{cimpoi2016deep} & $81.5_{\pm2.0}$ & $82.2_{\pm1.4}$ & $75.5_{\pm0.8}$ & $99.9_{\pm0.1}$ & $99.9_{\pm0.1}$ & - \\
				LFV \cite{song2017locally} & $82.6_{\pm2.6}$ & $82.1_{\pm1.9}$ & $73.8_{\pm1.0}$ & - & - & - \\
				DeepTEN \cite{zhang2017deep} & $82.0_{\pm{3.3}}$ & $80.2_{\pm0.9}$ & - & - & - & $84.3_{\pm1.9}$ \\
				Xception + SIFT-FV \cite{jbene2019fusion} & - & $86.1_{\pm1.6}$ & $75.4_{\pm1.0}$ & - & - & - \\
				DSRNet \cite{zhai2020deep} & $85.9_{\pm1.3}$ & $86.0_{\pm0.8}$ & $77.6_{\pm0.6}$ & - & - & $85.3_{\pm2.0}$ \\
				VisGraphNet \cite{florindo2021visgraphnet} & - & 77.3 & - & 98.1 & 97.6 & - \\
				Non-Add Entropy \cite{florindo2021using} & 84.4 & 77.7 & - & 98.8 & 98.5 & - \\
				Residual Pooling \cite{mao2021deep} & - & 85.7 & 76.6 & - & - \\
				FENet \cite{xu2021encoding} & $88.2_{\pm0.2}$ & $86.7_{\pm0.1}$ & $74.2_{\pm0.1}$ & - & - & $85.7_{\pm0.1}$ \\
				CLASSNet \cite{chen2021deep} & $87.7_{\pm1.3}$ & $86.2_{\pm0.9}$ & $74.0_{\pm0.5}$ & - & - & $85.6_{\pm2.2}$ \\
				DFAEN \cite{yang2022dfaen} & 86.6 & 87.6 & 76.1 & - & - & - \\
				RADAM \cite{scabini2023radam} & $90.7_{\pm4.0}$ & $88.7_{\pm0.4}$ & $77.0_{\pm0.7}$ & - & - & $84.2_{\pm1.7}$ \\ 
				Capsule \cite{mamidibathula2019texture} & 71.8 & 80.7 & 71.0 & - & 99.3 & \\
				\hline
				FCFVAE (Ours) & $92.3_{\pm4.6}$ & $92.4_{\pm0.2}$ & - & - & - & -\\
				FVAE (Ours) & $92.7_{\pm4.8}$ & $92.9_{\pm0.4}$ & - & - & - & -\\
				\hline
		\end{tabular}}
	\end{table}

	In summary, our results confirm the expectations that there are still many different ways of tackling self-supervised learning in computer vision. In particular, less computationally intensive architectures might be a promising direction to be further explored. Whilst models with large number of parameters, like vision transformers, yield remarkable performance in tasks where the relation among distant regions within the image play important role, this is not necessarily the case of other domains, such as the visual textures. Our study verify that ``U-Net-like'' architectures can be highly efficient in this scenario, providing rich deep representation of unlabeled images with relatively low computational cost.
	
	\section{Conclusions}
	\label{sec:conclusions}
	
	Here we proposed a self-supervised framework for texture recognition. Instead of relying on computationally intensive backbones, such as vision transformers, we adopt a convolutional simplified encoder-decoder module. The learnable features are processed by an advanced pooling operator inspired by Fisher vectors. Finally, the overall architecture is employed for texture recognition.
	
	Our proposal is evaluated on the classification of diverse texture datasets, achieving promising results. The accuracy outperforms several state-of-the-art models in the state-of-the-art with a low computational cost of the self-supervised module. The results suggest the potential still existent for convolutional architectures in self-supervised frameworks. Whereas ViTs are a natural choice in scenarios where the amount of data for pretraining is large, when little data is available, more efficient architectures are recommended.
	
	This study also points to aspects that demand further future investigation. One of such possibilities is how other self-supervised paradigms, like contrastive learning, could be adapted to benefit from efficient CNN backbones. Another one is a study on how other models from the state-of-the-art in texture analysis might be benefited by a self-supervised pretraining stage. In conclusion, we believe that these research lines will not only address current limitations of the literature but also pave the way for innovative applications and methodological advancements in the field.
	
	\section*{Code availability}
	
	The code developed in this study is publicly available at \url{https://github.com/lolyra/medical}.
	
	\section*{Declaration of competing interest}
	
	J. B. F. reports equipment, drugs, or supplies was provided by State of Sao Paulo Research Foundation. J. B. F. reports financial support was provided by National Council for Scientific and Technological Development. L. O. L. reports financial support was provided by Coordination of Higher Education Personnel Improvement.
	
	\section*{Acknowledgements}
	
	This study was financed in part by the Coordenação de Aperfeiçoamento de Pessoal de Nível Superior - Brasil (CAPES) - Finance Code 001.
	J. B. F. gratefully acknowledges the financial support of S\~ao Paulo Research Foundation (FAPESP) (Grant \#2020/01984-8) and of 
	National Council for Scientific and Technological Development, Brazil (CNPq) (Grant \#306981/2022-0).


\end{document}